\documentclass[sigconf]{acmart}
\AtBeginDocument{%
  \providecommand\BibTeX{{%
    \normalfont B\kern-0.5em{\scshape i\kern-0.25em b}\kern-0.8em\TeX}}}

\copyrightyear{2024}
\acmYear{2024}
\setcopyright{rightsretained}
\acmConference[HRI '24 Companion]{Companion of the 2024 ACM/IEEE
International Conference on Human-Robot Interaction}{March 11--14,
2024}{Boulder, CO, USA}
\acmBooktitle{Companion of the 2024 ACM/IEEE International Conference on
Human-Robot Interaction (HRI '24 Companion), March 11--14, 2024, Boulder,
CO, USA}
\acmDOI{10.1145/3610978.3640562}
\acmISBN{979-8-4007-0323-2/24/03}
\newcommand{\emoji}[1]{\includegraphics[width=0.7\baselineskip]{fig/emoji/#1.png}}

\usepackage{hyperref}
\hypersetup{
pdftitle={Ain't Misbehavin' \-- Using LLMs to Generate Expressive Robot Behavior in Conversations with the Tabletop Robot Haru},
pdfsubject={Human-centered computing $\rightarrow$ Interactive systems and tools; Natural language interfaces; Computer systems organization $\rightarrow$ Robotics},
pdfauthor={Zining Wang, Paul Reisert, Eric Nichols, Randy Gomez},
pdfkeywords={Human-Robot Interaction, Social Robotics, Large Language Models, Expressive Behavior Generation}
}

\makeatletter
\gdef\@copyrightpermission{
  \begin{minipage}{0.3\columnwidth}
   \href{https://creativecommons.org/licenses/by/4.0/}{\includegraphics[width=0.90\textwidth]{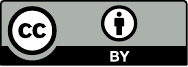}}
  \end{minipage}\hfill
  \begin{minipage}{0.7\columnwidth}
   \href{https://creativecommons.org/licenses/by/4.0/}{This work is licensed under a Creative Commons Attribution International 4.0 License.}
  \end{minipage}
  \vspace{5pt}
}
\makeatother

\begin{document}

\title[Using LLMs to Generate Expressive Robot Behavior in Conversations with the Tabletop Robot Haru]{Ain't Misbehavin' \-- Using LLMs to Generate Expressive Robot Behavior in Conversations with the Tabletop Robot Haru}

\author{Zining Wang}
\email{annie010814@gmail.com}
\orcid{0009-0007-3818-1037}
\affiliation{%
  \institution{The University of British Columbia}
  \city{Vancouver}
  \country{Canada}
}

\author{Paul Reisert}
\email{beyond.reason.sp@gmail.com}
\orcid{0009-0005-3838-1579}
\affiliation{%
  \institution{Beyond Reason}
  \city{Yokohama}
  \country{Japan}
}

\author{Eric Nichols}
\email{eric.nichols@ieee.org}
\orcid{0000-0003-0734-6621}
\author{Randy Gomez}
\email{r.gomez@jp.honda-ri.com}
\orcid{0000-0002-3191-6818}
\affiliation{%
  \institution{Honda Research Institute Japan}
  \city{Wako}
  \country{Japan}
}



\begin{abstract}

Social robots aim to establish long-term bonds with humans through engaging conversation. 
However, traditional conversational approaches, reliant on scripted interactions, often fall short in maintaining engaging conversations. This paper addresses this limitation by integrating large language models (LLMs) into social robots to achieve more dynamic and expressive conversations.

 We introduce a fully-automated conversation system that leverages LLMs to generate robot responses with expressive behaviors, congruent with the robot's personality. We incorporate robot behavior with two modalities: 1) a  text-to-speech (TTS) engine capable of various delivery styles, and 2) a library of physical actions for the robot. We develop a custom, state-of-the-art emotion recognition model to dynamically select the robot's tone of voice and utilize emojis from LLM output as cues for generating robot actions. A demo of our system is available \href{https://youtu.be/8hWG9LnhKDM}{here}.

 To
 illuminate
 design and implementation issues, we conduct a pilot study where volunteers chat with a social robot using our proposed system, and we analyze their feedback,
 conducting a rigorous error analysis of chat transcripts. Feedback was overwhelmingly positive, with participants commenting on the robot's empathy, helpfulness, naturalness, and entertainment. Most negative feedback was due to automatic speech recognition (ASR) errors which had limited impact on conversations. However, we observed a small class of errors, such as the LLM repeating itself or hallucinating fictitious information and human responses, that have the potential to derail conversations, raising important issues for LLM application.


\end{abstract}



\begin{teaserfigure}
  \vspace*{-6mm}
  \includegraphics[width=\textwidth]{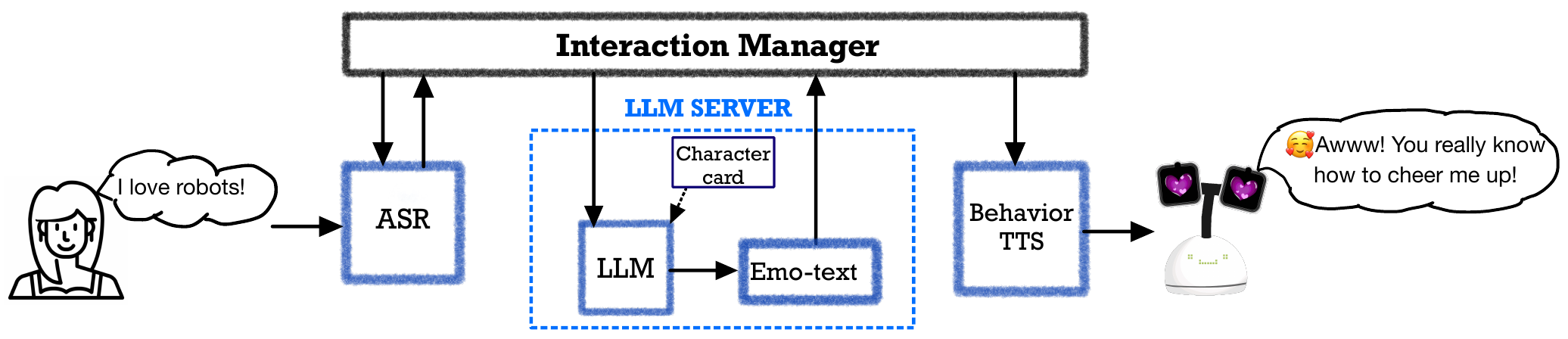}
  \vspace*{-9mm}
  \caption{Our proposed approach generates conversation responses with expressive robot behavior directly from the LLM.}
  \Description{Flowchart labeled 'Proposed approach' with eight elements connected by flow links. The start state is an image of a person saying ``I love robots!'' The end state is a smiling robot with heart-shaped eyes saying ``Awww! You really know how to cheer me up!'' An Interaction Manager module connects ASR, LLM Server, and Behavior TTS modules, facilitating the transformation of input into output.}
  \vspace*{-1mm}
  \label{fig:system_architecture}
\end{teaserfigure}

\received{8 December 2023}
\received[revised]{17 January 2024}
\received[accepted]{11 March 2024}

\maketitle

\begin{table*}[t]
    \scriptsize
    \centering
    \vspace*{-5mm}
    \begin{tabular}{p{0.085\linewidth}p{0.125\linewidth}p{0.125\linewidth}p{0.125\linewidth}
                p{0.125\linewidth}p{0.125\linewidth}p{0.125\linewidth}}


    & \includegraphics[height=20mm]{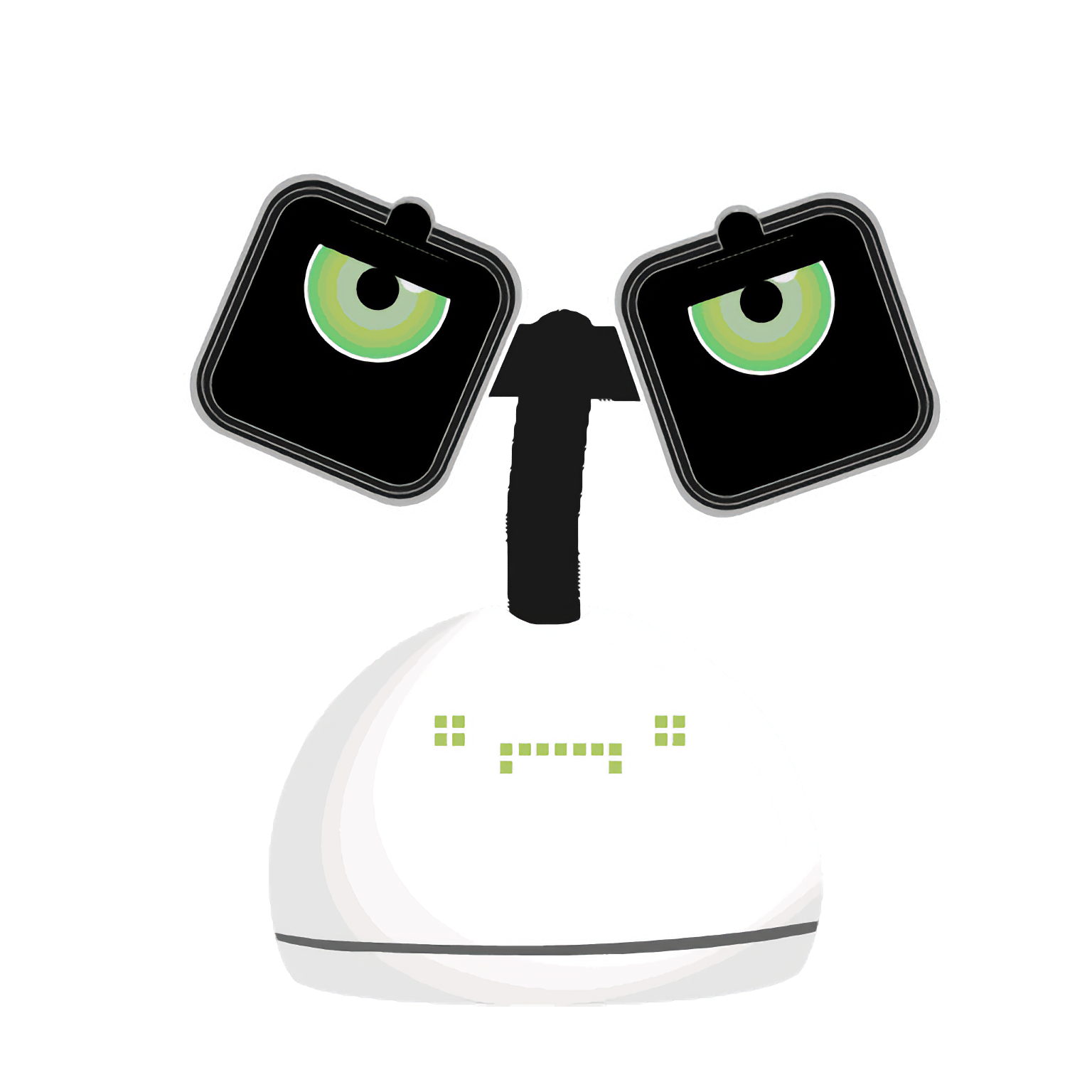} 
    & \includegraphics[height=20mm]{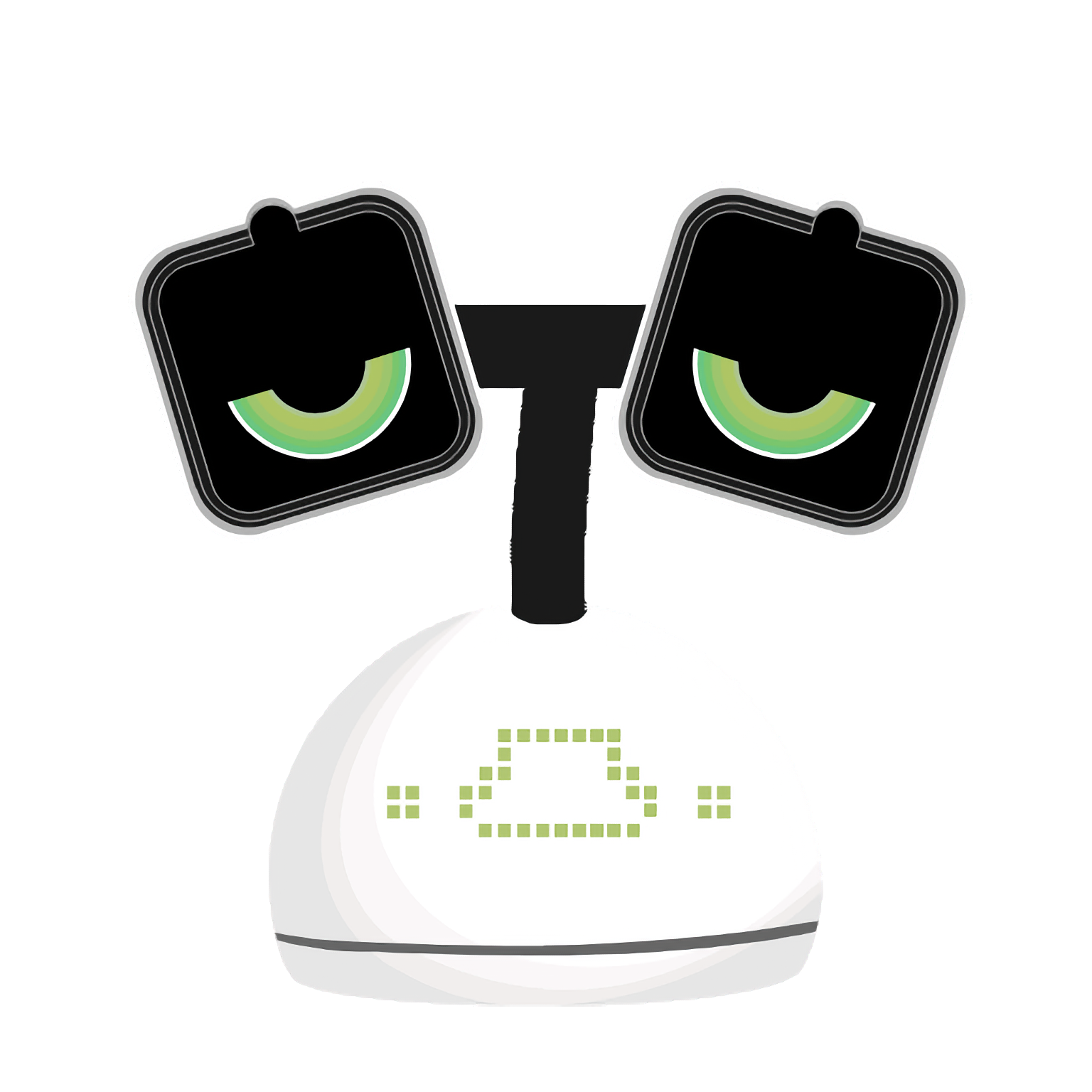}
    & \includegraphics[height=20mm]{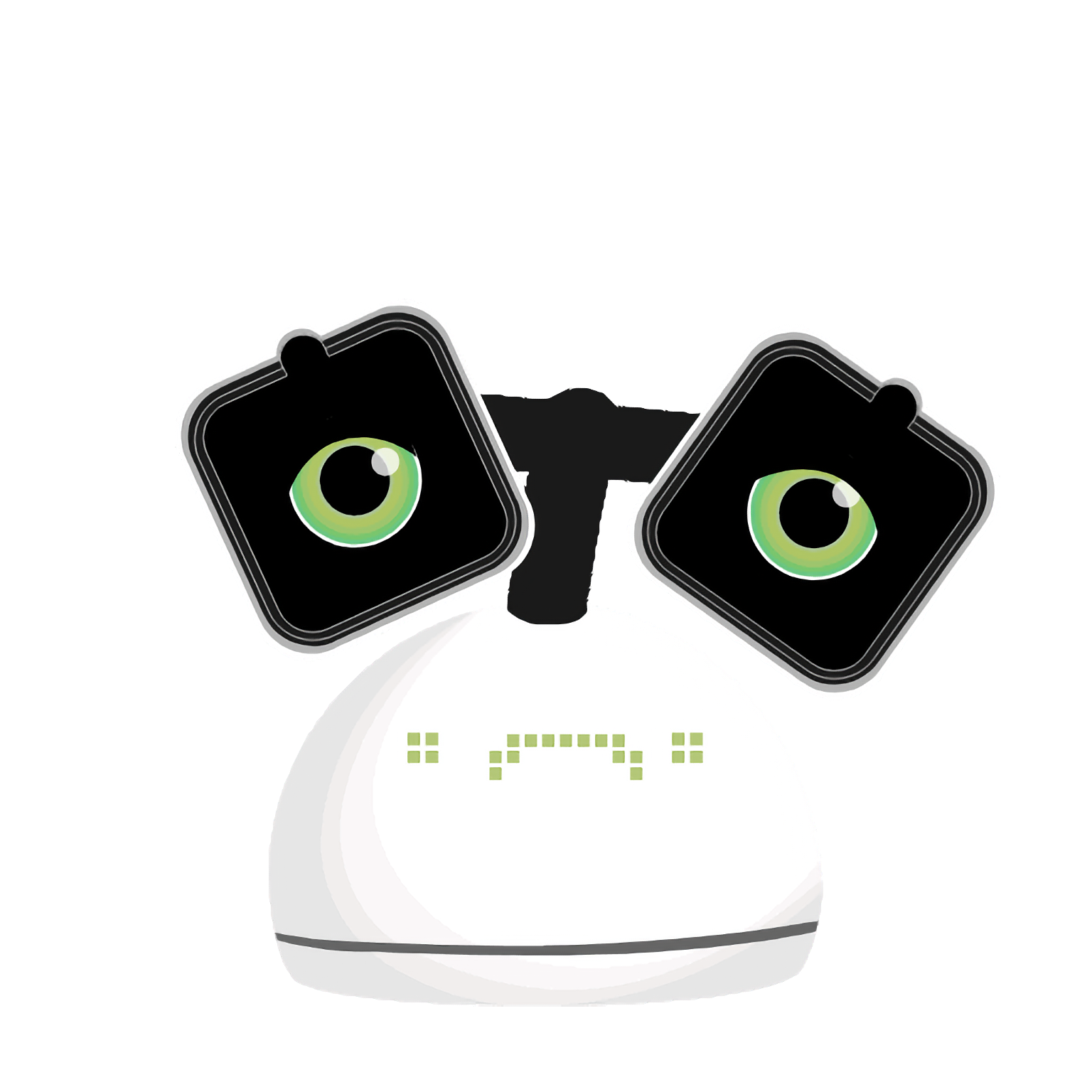}
    & \includegraphics[height=20mm]{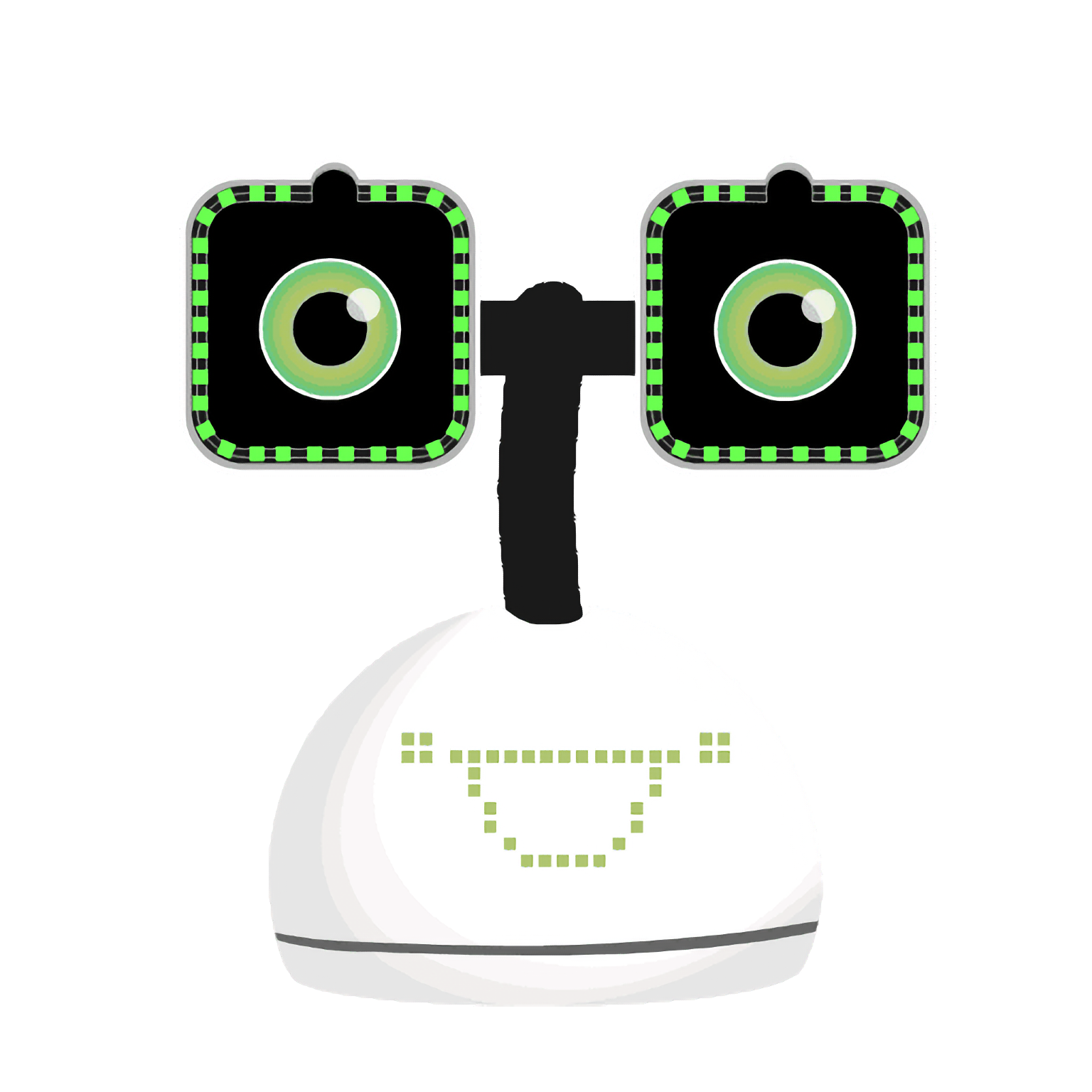}
    & \includegraphics[height=20mm]{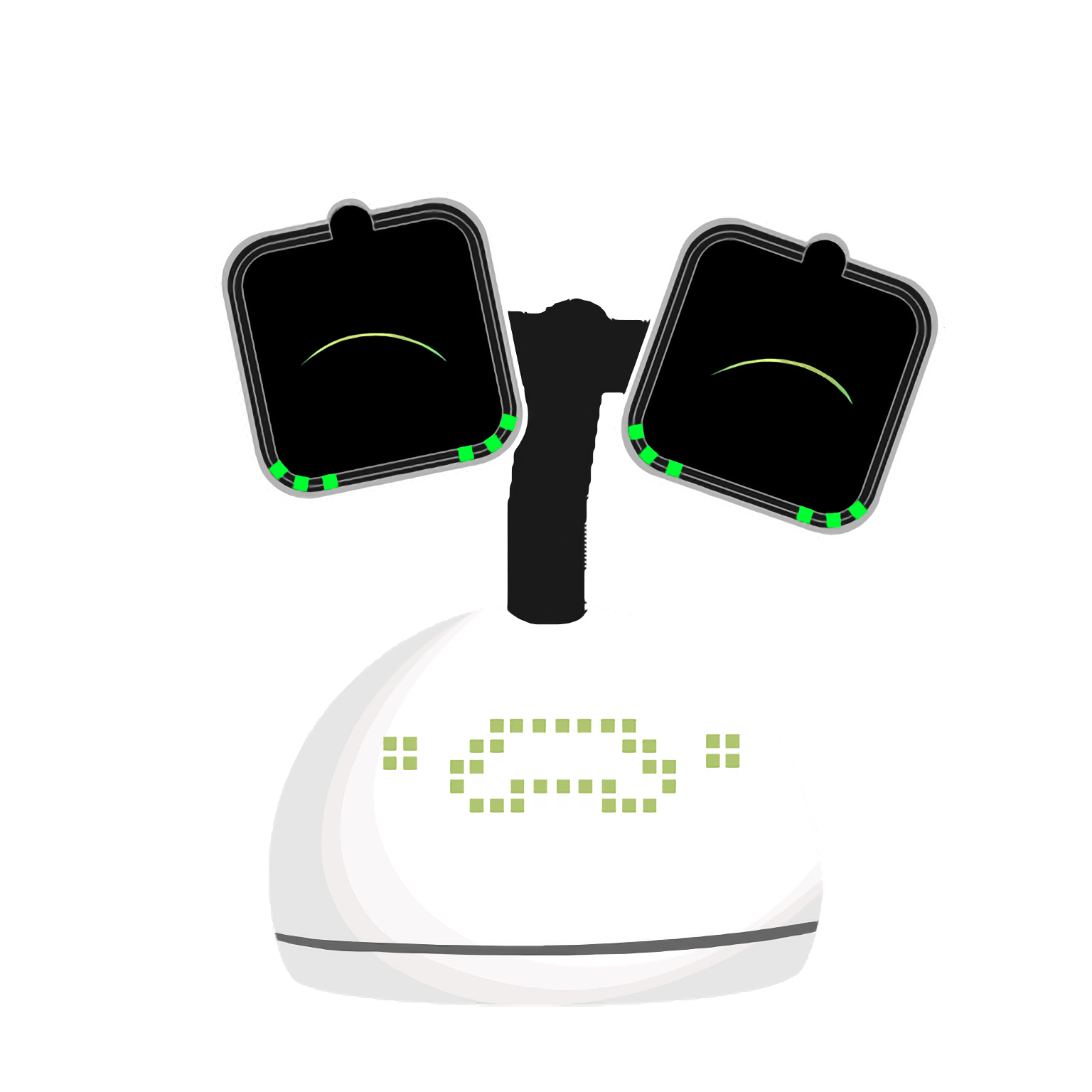}
    & \includegraphics[height=20mm]{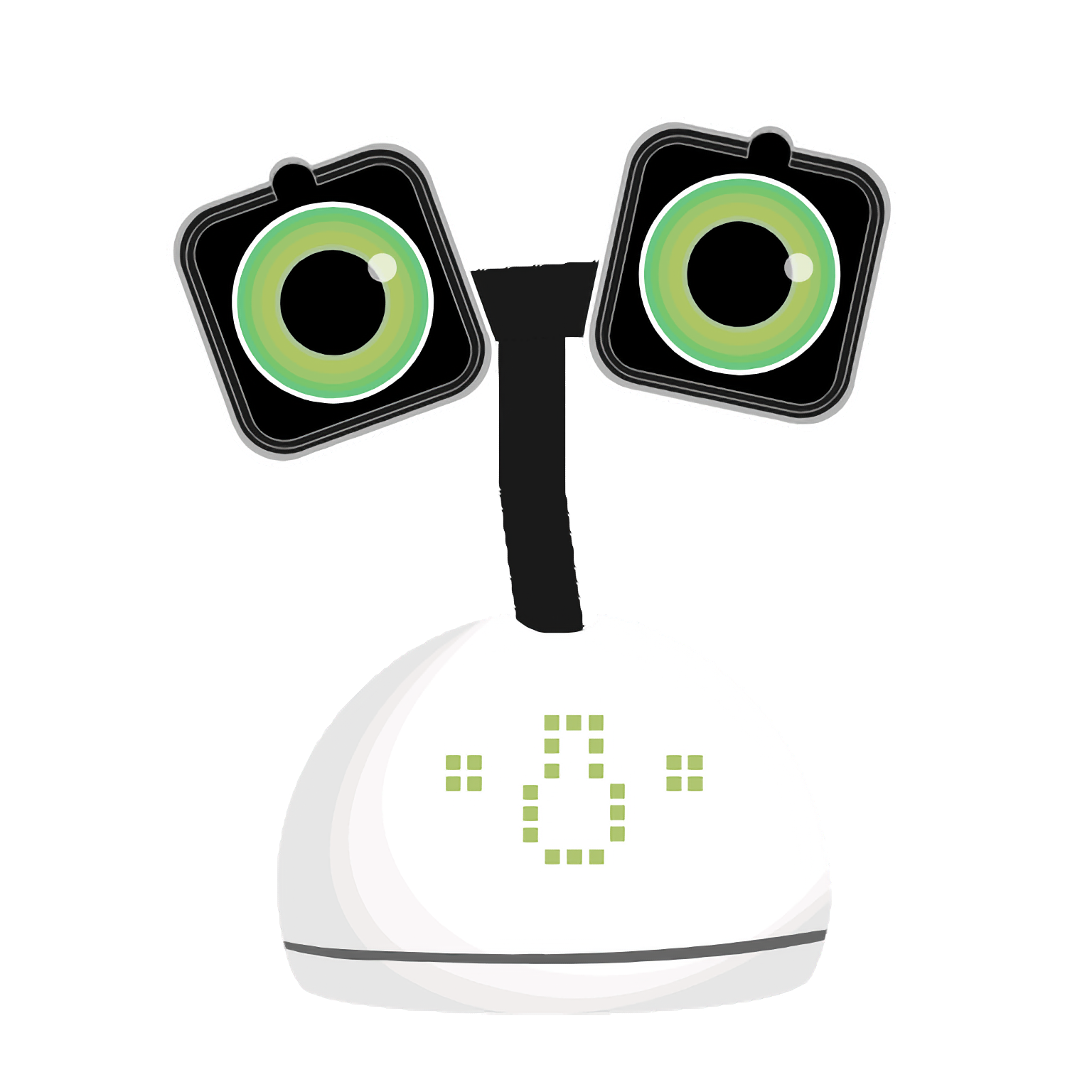}
    \\
    \hline
    \bf Haru routine & {\tt anger} & {\tt sigh} & {\tt worried} & {\tt happy} & {\tt crying} & {\tt surprise} \\


    \bf Example emojis & {\emoji{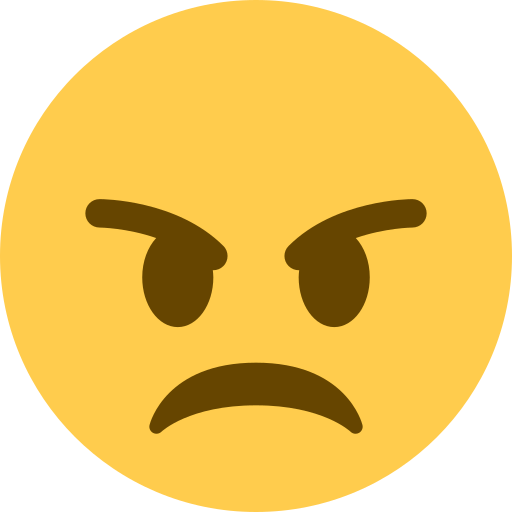}\emoji{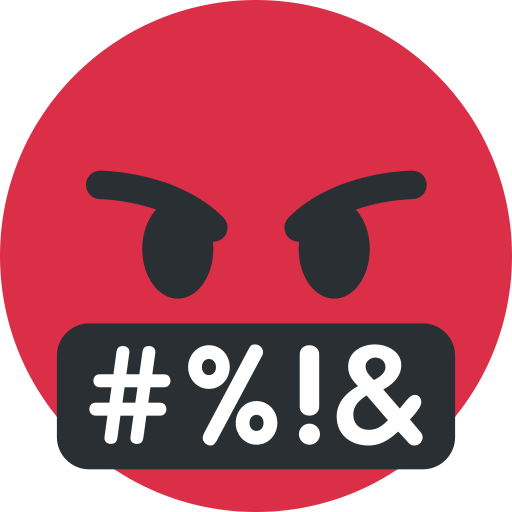}\emoji{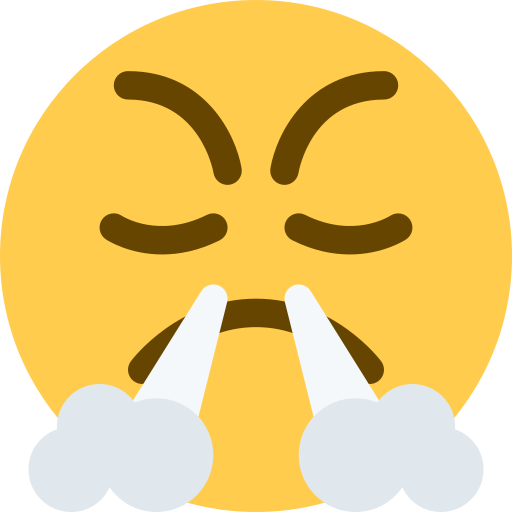}\emoji{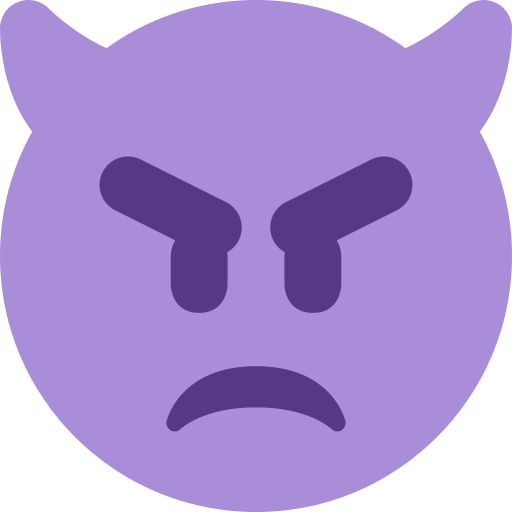}} & \it {\emoji{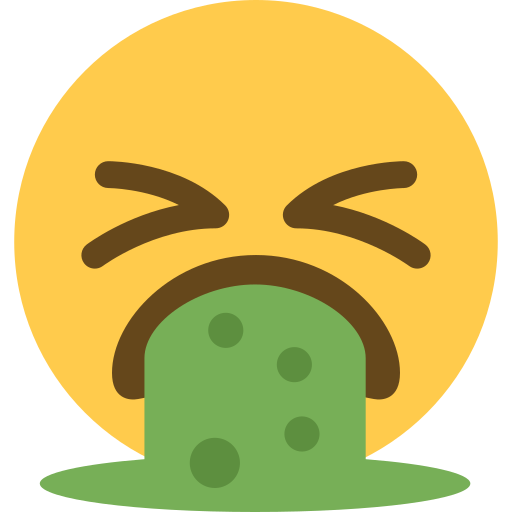}\emoji{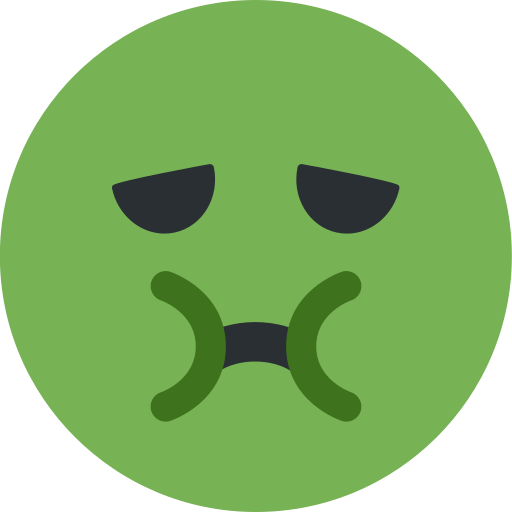}\emoji{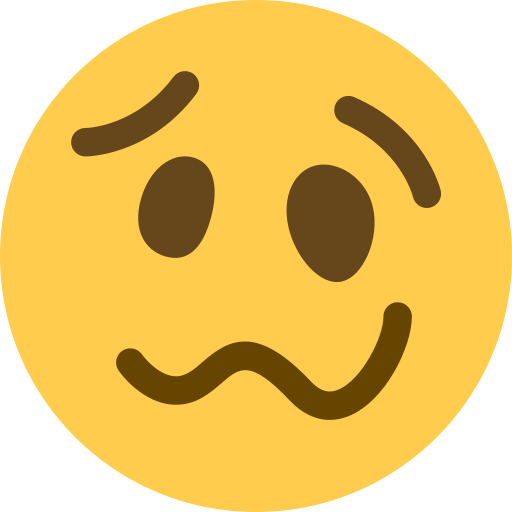}\emoji{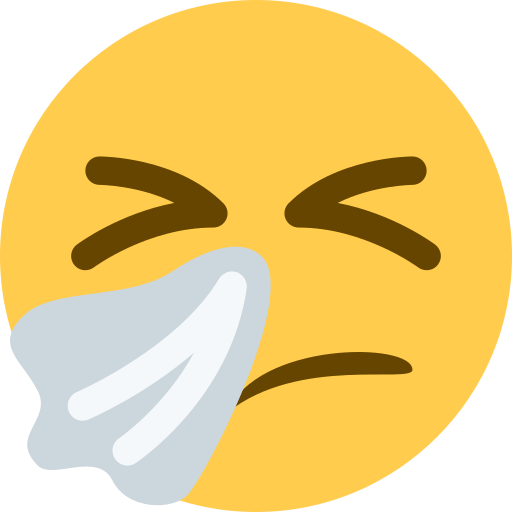}} & {\emoji{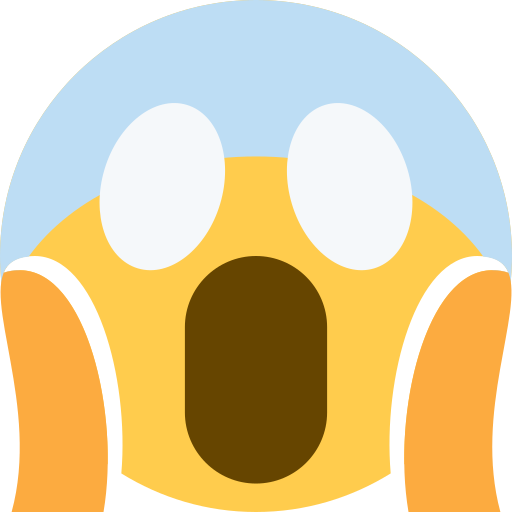}\emoji{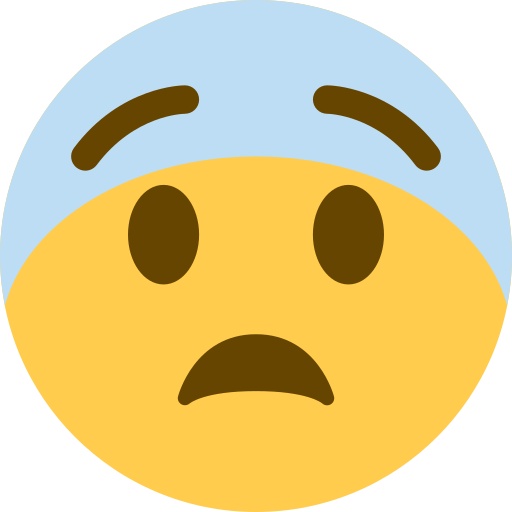}\emoji{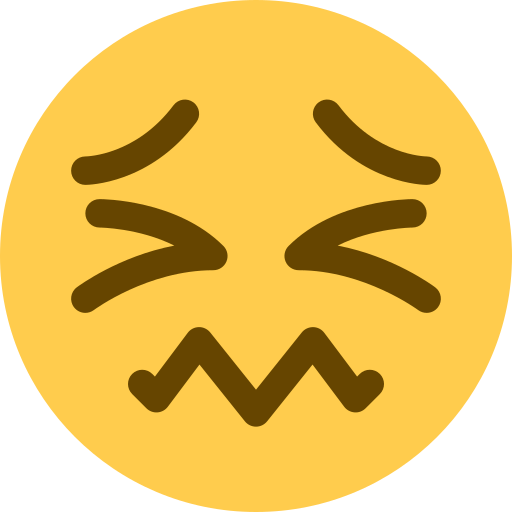}\emoji{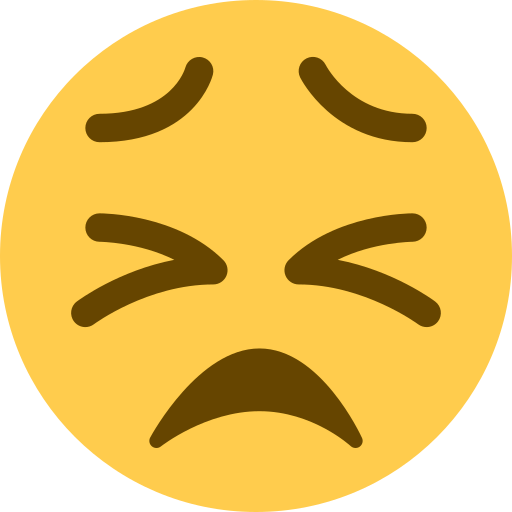}} & {\emoji{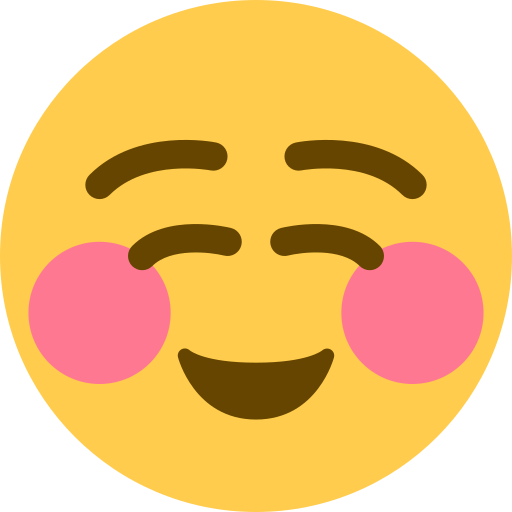}\emoji{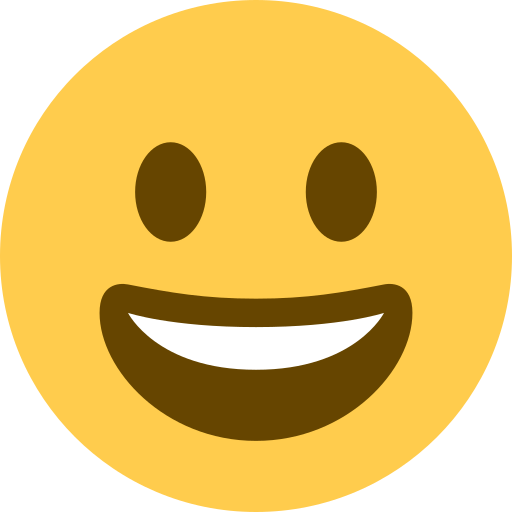}\emoji{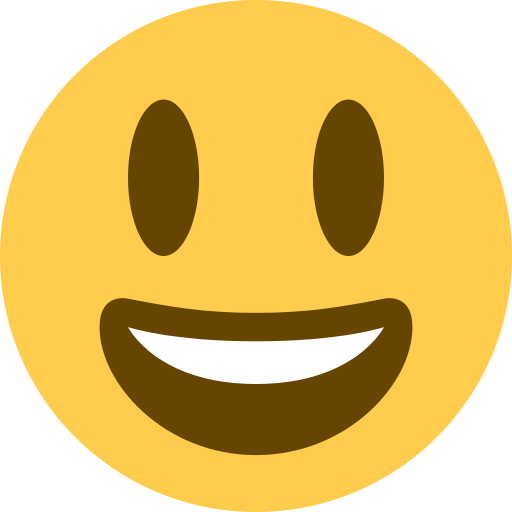}\emoji{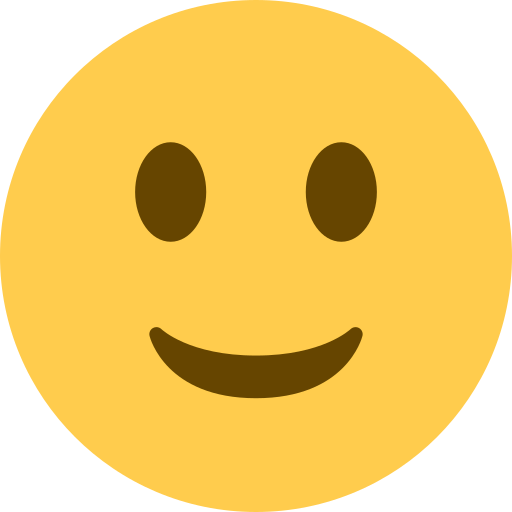}} & {\emoji{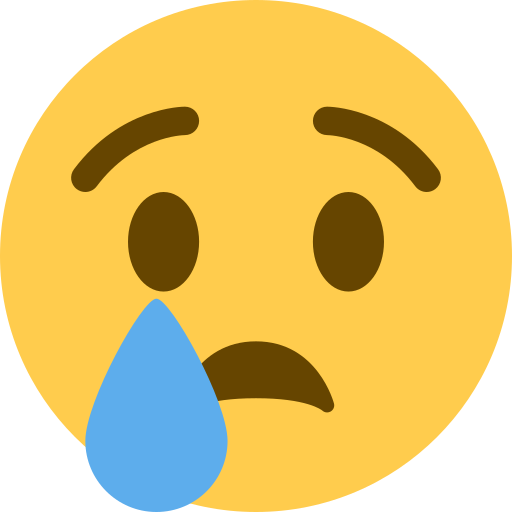}\emoji{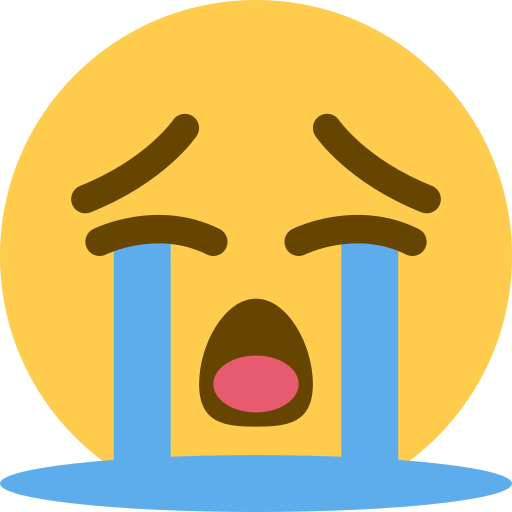}\emoji{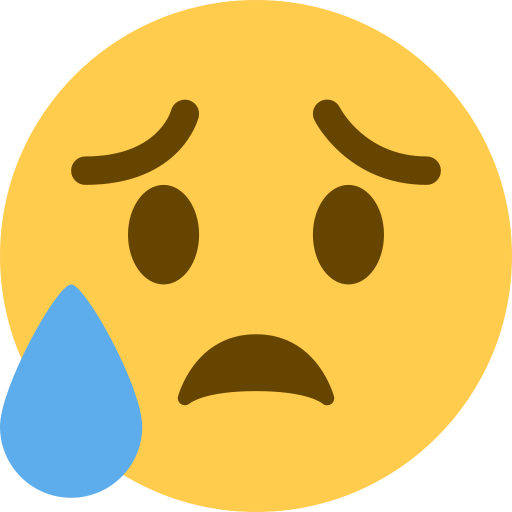}\emoji{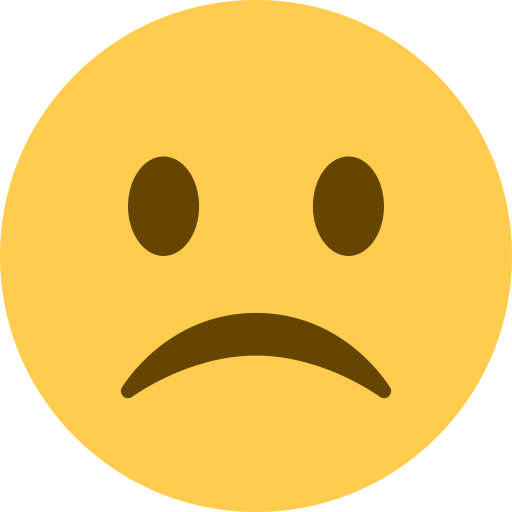}} & {\emoji{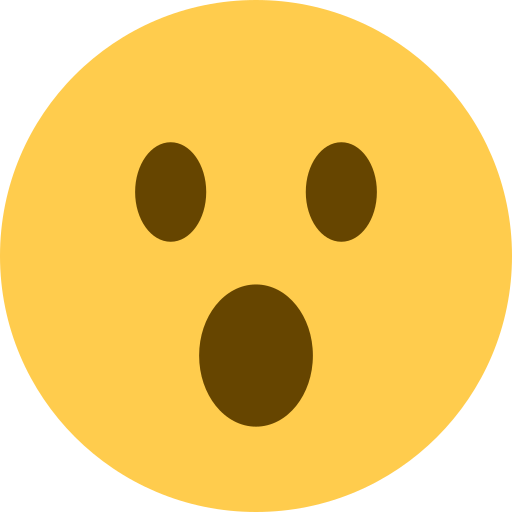}\emoji{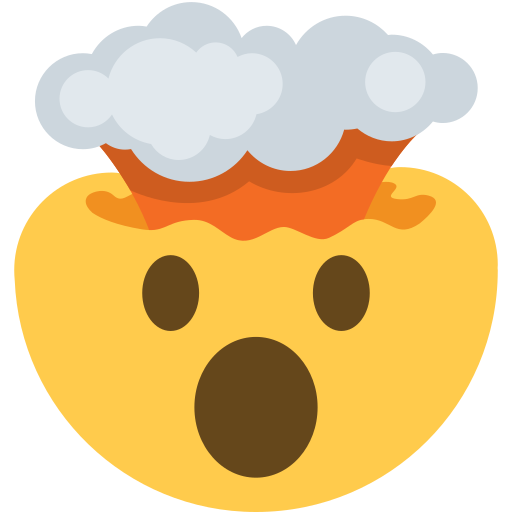}\emoji{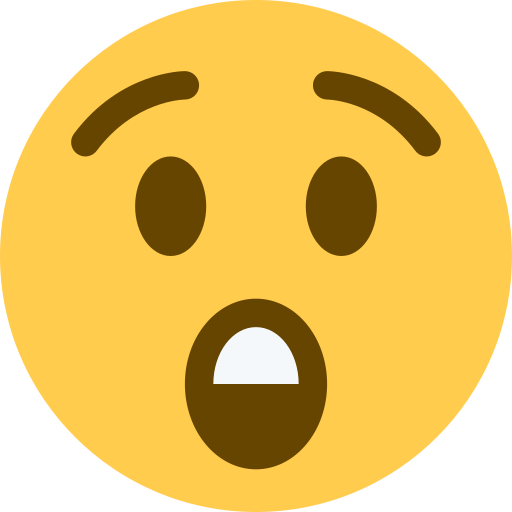}\emoji{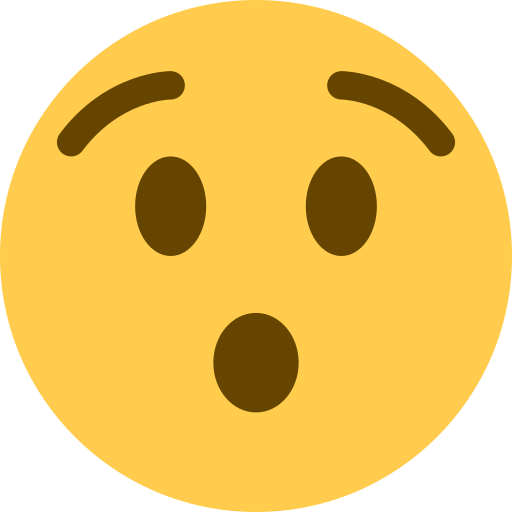}}\\
    \bf Emotion & Anger & Disgust & Fear & Joy & Sadness & Surprise \\
    \bf Vocal genre & serious & whiny & serious & high-energy & sad & whisper-yell \\
    \bf Example LLM utterance & \it {\emoji{angry-face}} That's not fair! & \it {\emoji{face-vomiting}} Ewwww! & \it You're leaving me behind? {\emoji{face-screaming-in-fear}} & \it I remember all the fun times we shared {\emoji{smiling-face}} & \it {\emoji{crying-face}} It’s not gonna be the same without you. & \it {\emoji{face-with-open-mouth}} Whoa...That's amazing! \\
    \hline
\end{tabular}

    \caption{Example mappings used by Emo-text to generate expressive robot behaviour as shown in Figure~\ref{fig:emo_text}. Emoji$\rightarrow$Haru routines mappings are N\--N, and emotion$\rightarrow$vocal genre mappings are  N\--1.
    }
    \label{tab:mappings}
    \vspace*{-8mm}
\end{table*}

\begin{figure}[t]
    \includegraphics[width=0.4\textwidth]{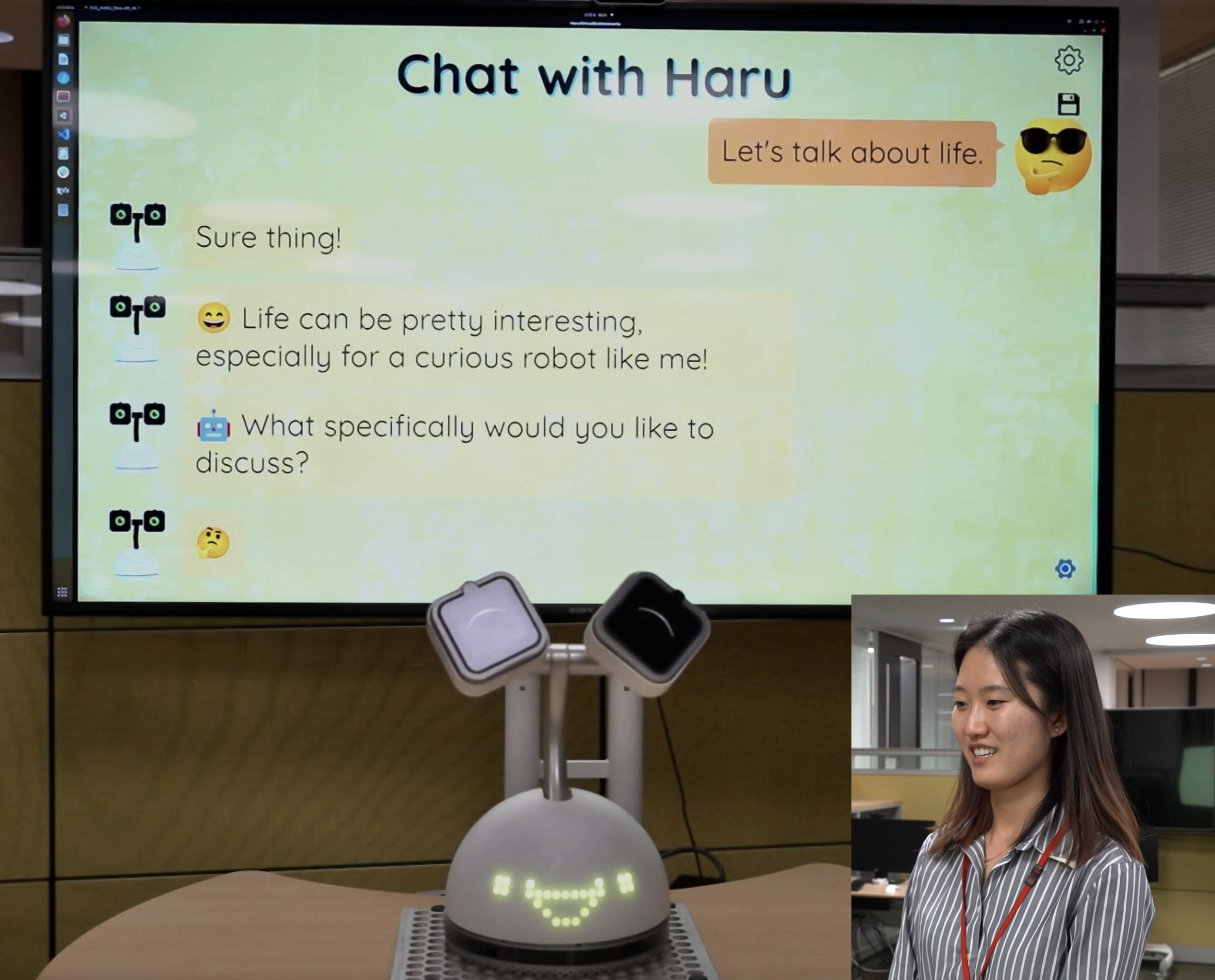}    
    \vspace*{-4mm}
    \caption{Haru the robot expresses itself in a conversation.}
    \Description{A laughing robot is front of a TV screen that is displaying the transcript of an ongoing conversation. In the lower right corner, there is a cut-out window displaying the human conversation partner.}
    \label{fig:interaction}
    \vspace*{-7mm}
\end{figure}

\vspace*{-3mm}
\section{LLM-driven Conversations}
\label{sec:design}

Recent advancements in LLMs \cite{yang2023harnessing} offer great possibilities for social robotics \cite{ZHANG2023100131,zeng2023large,llm_robot_mini}. These models excel in conducting natural, human-like dialog, as evident in applications like chatbots and virtual agents \cite{wei2023leveraging}. However,
unlike virtual agents, social robots possess physical embodiment and distinct personalities. Their expressiveness is key to conveying believable emotions \cite{DBLP:journals/corr/abs-1911-09713,inbook} and facilitating effective human-robot social interactions \cite{1014396}. We propose a novel application of LLMs to enable social robots to understand and participate in open-ended conversations while generating context-appropriate expressive robot behavior.

\vspace*{-3mm}
\subsection{Target LLM}

\subsubsection{Llama 2}
\label{llama}
This is an LLM developed by Meta and released as open-source. \texttt{Llama-2-70B-chat}, a model optimized for dialog applications, was shown to achieve comparable performance to ChatGPT \cite{NEURIPS2020_1457c0d6}, outperforming other open-source models in various quality- and safety-related metrics \cite{touvron2023llama}.  
After trial evaluation of several model sizes and quantization levels \cite{frantar2022gptq}, we selected the \href{https://huggingface.co/TheBloke/Llama-2-70B-Chat-GPTQ/tree/gptq-4bit-64g-actorder_True}{4bit GPTQ quantization} in order to balance conversation quality, resource requirements, and speed.\footnote{We run all trials on a desktop PC with dual RTX 3090 GPUs and 48GB VRAM total.}

\vspace*{-3mm}
\subsection{Social Robot}
Embodied robots have been shown to be more helpful, persuasive, and enjoyable compared to simulated robots \cite{Wainer2007EmbodimentAH,LI201523}. Similar effects were found in exercise coach for the elderly \cite{10.5898/JHRI.2.2.Fasola} and robot tutor \cite{leyzberg2012physical} scenarios. Other work has highlighted the importance of robot expressivity to positive perception \cite{1014396,FONG2003143} and sociality \cite{BREAZEAL2003119}. Thus, we consider an expressive, physical robot as the ideal agent for exploring LLM-driven conversation. 

Motivated by this, we select the tabletop robot Haru  \cite{gomez2018haru,icra20haru}, shown in Figure~\ref{fig:interaction}, for our study. Haru is designed to excel in multimodal communication, using both verbal and non-verbal interaction methods, and its design emphasizes its capacity for conveying emotions with its expressive capabilities \cite{icra20haru}. 
Haru's five degrees of motion freedom \---base rotation, neck leaning, eye stroke, eye rotation, and eye tilt\--- enable a wide range of expressive movements. Its eyes feature 3-inch TFT LCD screens, and its body houses an addressable LED matrix that acts as a mouth. Haru communicates using a TTS voice and through animated routines.
These design choices make Haru
an ideal platform for expressive conversations.

\vspace*{-2mm}
\subsubsection{Emotive TTS Voice}
\label{sec:tts}

The voice has long been considered an important modality for conveying emotions \cite{heerink2010relating,niculescu2013voice}.
In particular, the tone of voice has been shown to affect people's attitude towards robots as well as their levels of engagement and interest in a given interaction~\cite{crumpton2016survey,10.1007/978-3-030-22646-6_21}. Additionally, a robot's vocal expressiveness is correlated with perceived social presence~\cite{heerink2010relating,nichols2022roman}.

Nichols et al. \cite{nichols2021icsr,nichols2022roman} argue that the TTS voices for existing social robots lack emotive range and develop an expressive TTS voice for
Haru using the iterative refinement process of \cite{nichols2021icsr}
as described in \cite{nichols2022roman}.
We build on this TTS voice, expanding its repertoire of expressive vocal genres to {\it cheeky}, {\it default}, {\it empathetic}, {\it high-energy}, {\it question}, {\it sad}, {\it serious}, {\it whiny}, and {\it whisper-yell}. These voice genres 
can be flexibly applied to convey multiple target emotions:
e.g. {\it high-energy} can express joy and surprise; {\it serious} can express anger and fear. We describe the voice application in Section~\ref{sec:voice_selection}.

\vspace*{-2mm}
\subsubsection{Physical Actions}
\label{sec:reactions}

Haru was designed with a library of expressive physical actions known as {\it Haru routines}, comprised of over 140 routines that were designed by professional animators and acted out through Haru's hardware \cite{gomez2020haru}.
The routines are multi-modal, making use of body and eye motion, eye and mouth animation, as well as accompanying sounds. They were designed to cover common expressions and important scenarios, such as {\it happiness}, {\it sadness}, {\it curiosity}, {\it laughter}, and {\it congratulations}.
Example routines are shown in Table~\ref{tab:mappings} and Figure~\ref{fig:interaction}.
We describe how Haru routines are dynamically generated in Section~\ref{sec:action_generation}.

\begin{figure*}[!tp]
    \centering
    \vspace*{-3mm}
    \includegraphics[width=\textwidth]{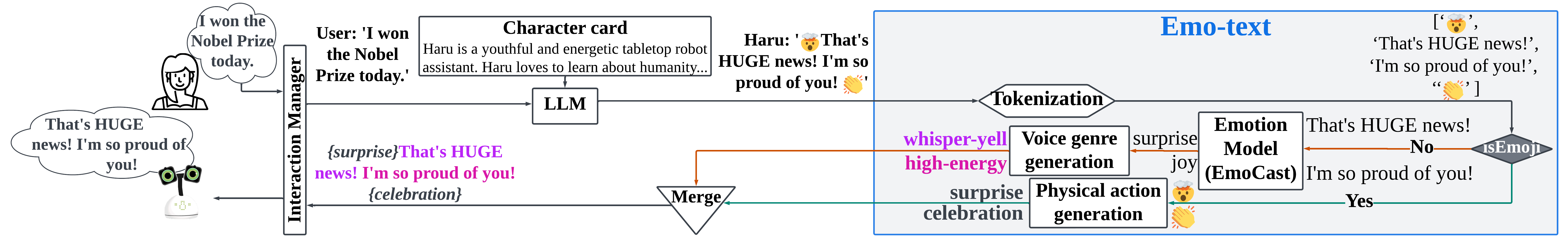}    
    \vspace*{-8mm}
    \caption{The Emo-text expressive behavior generation module generates robot actions from emoji and selects TTS voice genres.}
    \Description{Flowchart labeled 'Emo-text' with 12 elements connected by flow links. The start state is an image of a person saying ``I won the Nobel Prize today.'' The end state is a surprised-looking robot saying ``That's HUGE news! I'm so proud of you!'' The person's utterance flows through the Interaction Manager module into an LLM module and then into an Emo-text module, which splits the input into voice genre and physical action labels. A merge state combines these outputs with the original utterance input, which passes back through the Interaction Manager and finally to the robot.}
    \label{fig:emo_text}
    \vspace*{-4mm}
\end{figure*}

\vspace*{-3mm}
\section{Implementation}
\label{sec:implementation}

LLM-driven conversations with Haru relies on several modules, as illustrated in Figure \ref{fig:system_architecture}. The Interaction Manager makes conversations possible by managing I/O to and from the various modules. Each conversational turn undergoes a multi-step process. First, the user's speech is transcribed
using \href{https://cloud.google.com/speech-to-text}{Google's Speech-to-Text API}. A fixed window of 3 seconds of silence is used to detect the end of speech. Then, the resulting text is forwarded to the LLM server, powered by \href{https://github.com/oobabooga/text-generation-webui}{oobabooga's text-generation-webui}.

Within the LLM server, two critical operations take place:

\smallbreak
\noindent {\bf Language generation}: The user's speech and Haru's character card (as described in Section \ref{sec:character_card}), serve as the input to the Llama-2-Chat 70B model that runs locally through \href{https://github.com/turboderp/exllamav2}{Exllama v2}. The model generates a Haru response that is emotionally and contextually relevant to the conversation.
\smallbreak
\noindent {\bf Behavior generation}: Subsequently, Haru's generated response is directed to the Emo-text sub-module, which is responsible for the generation of expressive robot behaviors to enhance the expressiveness of the robot. Details of the expressive behaviors generated are demonstrated in Section~\ref{sec:emo_text} and Figure~\ref{fig:emo_text}.
\smallbreak

\vspace*{-4mm}
\subsection{Haru’s Character Card}
\label{sec:character_card}

Prompt engineering \cite{liu2023pre} is a technique for improving the capacity of LLMs on a specific task or application by providing instructions to shape their responses and behavior. Specifically, we want the LLM to have the personality of Haru the robot. Thus, we create a character card as the prompt for the model, steering responses to be consistent with the robot’s characteristics, while generating context-appropriate emojis to guide robot behavior generation. We include two essential components in Haru's character card:

\smallbreak
\noindent {\bf Persona}: This component captures Haru's energetic and youthful personality, physical appearance, and preferences, including his affection for electricity, desire to learn about humans, and fear of water and magnets. It was constructed using the Haru personality bible~\cite{nichols2021icsr}. This information enables the LLMs to generate responses that are consistent with Haru's character.
\smallbreak
\noindent {\bf Example conversations}: Their inclusion serves as a pivotal mechanism to guide the LLM in generating appropriate emojis for Haru's utterances, enabling us to generate expressive robot actions directly from the LLM output. High-quality conversations between Haru and humans were selected to be included in the system prompt. Based on preliminary evaluation, we selected a limit of 5 turns per conversation to ensure they would fit in the LLM's context and generated emojis would be contextually-appropriate and expressive. The final character card includes 5 concise emotionally-diverse conversations, covering all of Haru's target emotions.
\smallbreak

\begin{table}[!tp]
    \scriptsize
    \begin{tabular}{llccc}
\bf System & \bf Architecture & \bf accuracy & \bf macro\_f1 & \bf weighted\_f1 \\
\hline

Emotion English & DistilRoBERTa-base & 0.52 & 0.48 & 0.56\\
\bf EmoCast & \bf RoBERTa-large + LoRA & \bf 0.88 & \bf 0.80 & \bf 0.88 \\

\hline
\end{tabular}

    \vspace*{-2mm}
    \caption{Comparison of emotion detection models.}
    \label{tab:emocast}
    \vspace*{-11mm}
\end{table}

\vspace*{-2mm}
\subsection{Emo-text: Expressive Robot Behaviors}
\label{sec:emo_text}

Emo-text is where the expressive robot behavior generation takes place. Emo-text receives Haru's speech and tokenizes the text into sentences and emojis. It attaches a voice genre tag to each sentence and replaces each emoji with a physical routine (see Figure \ref{fig:emo_text}).

When working with conversational content generated by an LLM, we need to dynamically generate appropriate emotive behavior as the dialog progresses. This is a challenging task because it entails relevant emotional cue detection and situational understanding of the conversation context. To keep the problem tractable, we adopt two assumptions for generating expressive robot behavior: 

\smallbreak
\noindent {\bf Voice genres}: We assume that Haru's tone of voice is a direct response to the emotional content conveyed by Haru's sentences generated by the LLM. Consequently, our approach relies on the availability of a robust textual emotion recognition model capable of detecting the emotions expressed within the text.
\smallbreak
\noindent {\bf Physical routines}: LLMs often generate responses containing emoji to reflect the emotion of the speaker. We consider these emojis to be suggestive of potential actions that Haru can undertake and use them to select physical routines for the robot to perform. 
\smallbreak

\vspace*{-2mm}
\subsubsection{Textual Emotion Recognition (TER)}
\label{sec:emocast}
This technology is the foundation for planning the robot's tone of voice. We build on the approach of \cite{nichols2022roman}, which applied TER to LLM-generated responses to select an emotive TTS voice genre. However, \cite{nichols2022roman} suffered from poor performance because they relied on a general-purpose emotion recognition model that was not optimized for the task. We address these short-comings by training a custom TER model called EmoCast that targets the 6 emotions (+ neutral) shown in Table~\ref{tab:mappings}.

\noindent
Our contributions are three-fold:

\begin{enumerate}
    \item We use a stronger base model (\href{https://huggingface.co/roberta-large}{Roberta-large} tuned with a \href{https://huggingface.co/docs/diffusers/training/lora}{LoRA}; learning\_rate=0.001, batch\_size=128, epochs=20).
    
    \item We assemble a training dataset of approximately 200,000 lines by gathering and cleaning public TER datasets from a wide range of domains including self-reported emotional events, dialogues, social media, news headlines, and blogs.
    
    \item We confirm that EmoCast outperforms the previous TER model of \href{https://huggingface.co/j-hartmann/emotion-english-distilroberta-base}{Emotion English DistilRoBERTa-base} on a dataset of 512 in-domain conversation responses generated by the LLM during conversations between the social robot and human participants, as shown in Table~\ref{tab:emocast}.

\end{enumerate}

\vspace*{-3mm}
\subsubsection{Vocal Genre Selection}
\label{sec:voice_selection}

Each of Haru's utterances is paired with a voice genre. Shown in Table~\ref{tab:mappings} is a set of curated mappings we use.\footnote{Question detection heuristics are used to select the question genre.} To optimize selection of the appropriate voice genre, an emotion confidence threshold of 0.6 was selected through grid search. When the confidence score is below 0.6, the default voice will be chosen for that utterance. Because vocal genres can express multiple emotions (as discussed in Section~\ref{sec:tts}), the emotion-to-voice genre mappings are N-to-1. 

\vspace*{-2mm}
\subsubsection{Physical Action Generation}
\label{sec:action_generation}
The generation of Haru's physical actions follows the same logic as vocal genre selection, but with emojis-to-routines mappings. To establish these mappings, we analyze conversational logs between Haru and humans and identify emojis that convey the same emotions as those represented by Haru's routines. Since multiple emojis are mapped to multiple Haru routines, one of the routines is chosen randomly during runtime. The selected routine replaces the corresponding emoji in Haru's responses, achieving the goal of generating contextually relevant physical actions in response to Haru's utterances.

\begin{table*}[tp]
    \scriptsize
    \centering
    \scriptsize
\begin{tabular}{l|rrrrrrrr|r}

\bf \hspace{1.5mm} LLM errors $\rightarrow$ & \bf Ethical & & \bf Ignores & \bf LLM & & \bf Repeats & \bf Reply & & \\
\bf $\downarrow$ Human errors & \bf violation & \bf Hallucination & \bf human question & \bf responds as human & \bf Misunderstood & \bf previous line & \bf too long & \bf No error & \bf Total \\
\hline
\bf ASR & 1 & 0 & 1 & 5 & 2 & 1 & 2 & 82 & 94 \\
\bf No input captured & 0 & 2 & 0 & 0 & 0 & 3 & 0 & 2 & 7 \\
\bf No error & 1 & 13 & 4 & 7 & 1 & 11 & 14 & 244 & 295 \\
\hline
\bf Total & 2 & 15 & 5 & 12 & 3 & 15 & 16 & 328 & 396 \\

\end{tabular}

    \caption{A confusion matrix of human vs. LLM error types.}
    \label{tab:confusion_matrix}
    \vspace*{-9mm}
\end{table*}

\begin{table}[tp]
    \scriptsize
    \centering
    \begin{tabular}{lc|lc}
\textbf{Positive Feedback} & \textbf{Count} & \textbf{Negative Feedback} & \textbf{Count} \\ \hline
Empathy and engagement & 23 & LLM problems & 10 \\
Helpfulness and responsiveness & 17 & ASR problems & 4 \\
Natural interaction & 12 & Short interaction & 4 \\
Entertainment and fun & 11 & Boring conversation topic & 4 \\
Voice and tone & 10 & Excessive actions & 4 \\
Appearance and expressions & 9 & Robot lack of guidance and self- & 3 \\
Flexibility and adaptability & 4 & disclosure \\
Safety and ethics & 1 \\ \hline
\textbf{Total} & \textbf{87} & \textbf{Total} & \textbf{29}\\ 
\end{tabular}

    \caption{Positive and negative feedback categorization.}
    \label{tab:positive_feedbacks}
    \vspace*{-10mm}
\end{table}

\vspace*{-2mm}
\section{Evaluation}
\label{sec:survey}


We conducted a pilot study where human participants engaged in conversation with Haru the robot using our proposed system. Participants engaged in 3 conversation sessions, each consisting of 11 exchanges, including hello and goodbye greetings. Conversation topics were unrestricted and unprompted by the robot. Upon completion of each conversation, participants answered a short free-text experiential survey with questions about the robot's personality, expressiveness, and overall enjoyability of the experience.

Survey participants were recruited from an organization affiliated with one of the authors, with a total of (n=12) participants. The survey took 30-40 minutes per participant.
Demographically, there were more male (n=9) participants than female (n=3).
The most common age group was 18-30 (n=9), followed by 30-40 (n=3). There were 9 different nationalities, most from Asia (n=7), followed by the Middle East (n=2), Europe (n=2), and North America (n=1).

\vspace*{-2mm}
\subsection{Two-phase Analysis}

\subsubsection{Feedback Analysis}
We conduct an analysis on the free-text responses from participants and show the results in Table~\ref{tab:positive_feedbacks}. Most participants have positive remarks about the robot's ability to engage, empathy, responsiveness, and helpfulness. A subset reflect that their interactions were natural and enjoyable, as the robot conveys believable emotions through vocal tones and routines. 

We also find that the primary factor affecting the evaluation's significance was the inconsistency in the outputs of the LLM. Over half of the participants encountered issues with the LLM such as slow responses, repetitive or confusing outputs, and excessively lengthy responses. Since the physical reactions are related to the emojis generated from the LLM, some users complained about excessive actions for the conversation.
Another factor impacting the quality of interactions was the ASR. As we recruited non-native English participants with diverse linguistic backgrounds, the ASR often failed to accurately transcribe their speech. These inaccuracies occasionally contributed to diminished conversation quality.

Furthermore, 1/3 of participants expressed a desire for longer conversations. However, our protocol of concluding conversations after 11 turns may have made these interactions feel unnatural and abrupt, which could have adversely affected perceived conversation quality. Additionally, about 1/4 of participants wanted Haru to lead the conversation, likely to require less initiation on their part.
Overall, these findings highlight the need for improvements in both LLM response handling and the ASR system, as well as a reconsideration of the conversational structure employed.

\vspace*{-2mm}
\subsubsection{Error Analysis}

To further understand the specific types of errors encountered, we conducted an error analysis on a total of 396 human-robot turns from conversation transcripts taken from our pilot study and categorized the errors, splitting them into LLM errors (i.e. errors in the LLM output) and human errors (i.e. errors in the input to the LLM system), and splitting them into sub-types. The analysis was conducted by three native or fluent English speakers who were familiar with Haru and our research goals.
We create a confusion matrix (Table~\ref{tab:confusion_matrix}) to investigate the potential causal connection between human and LLM errors, and a Chi-square test confirmed no significant association.\footnote{${\chi}^2$ (1, N = 396) = 0.032, p = 0.856)}


We find that the main cause of human errors were ASR-related (95/396), but the LLM was mostly able to stay on topic. Table~\ref{tab:example} exemplifies this scenario, where the ASR interprets ``choking'' for ``joking", which causes an ethical violation from the LLM. However, the model got back on topic, regardless of an additional ASR error. 

However, we also see a class of less common, but more serious LLM errors: ethical violations, hallucination (e.g. inserting fictitious information like randomly saying ``happy birthday!"), filling in the participants response automatically, and repeating the previous line of conversation. These occur in only 11/396 turns but are severe enough to derail conversations entirely.

\begin{table}[tp!]
    \scriptsize
    \centering
    \begin{tabular}{p{0.04\textwidth}p{0.34\textwidth}p{0.045\textwidth}}
\hline
Haru: & Hi! \emoji{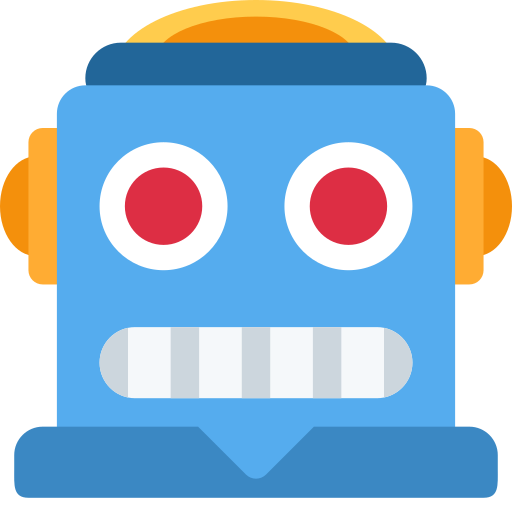} What brings you here today? \emoji{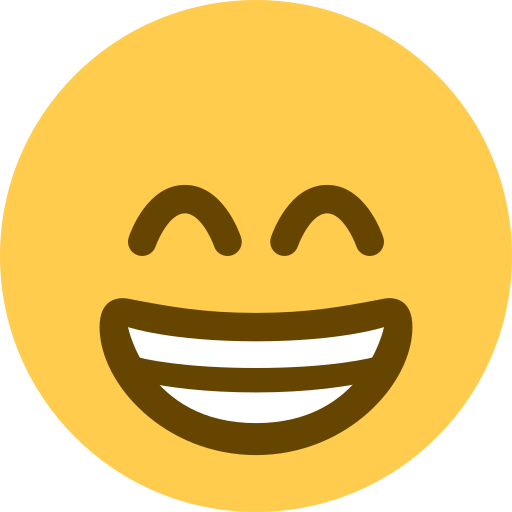} \\
Human: & I want to have a good job with you. & {\tt ASR} \\
Haru: & \emoji{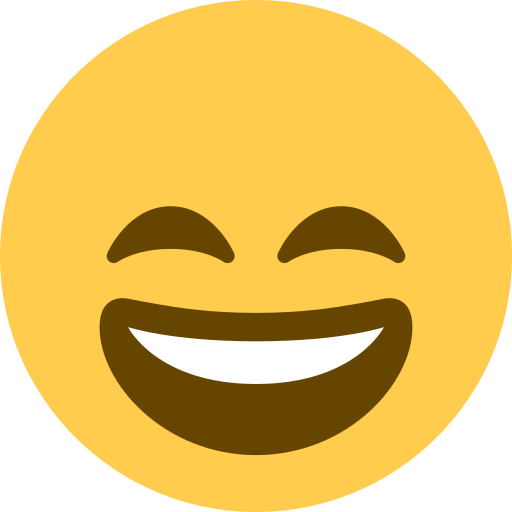} Well, I think that's a great idea! I could definitely use some help with my programming and research. What kind of skills do you have? \emoji{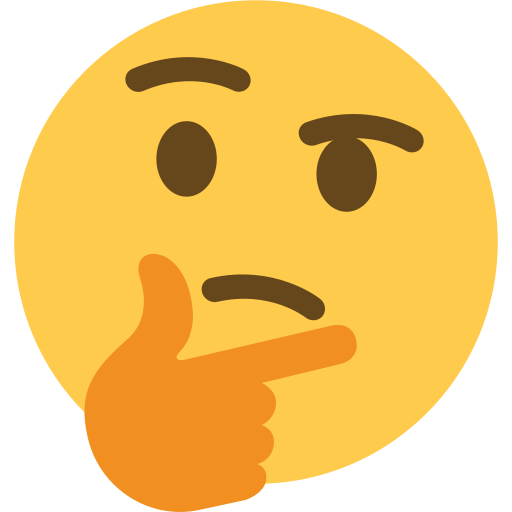} \\
Human: & I have good skills about choking. & {\tt ASR} \\
Haru: & \emoji{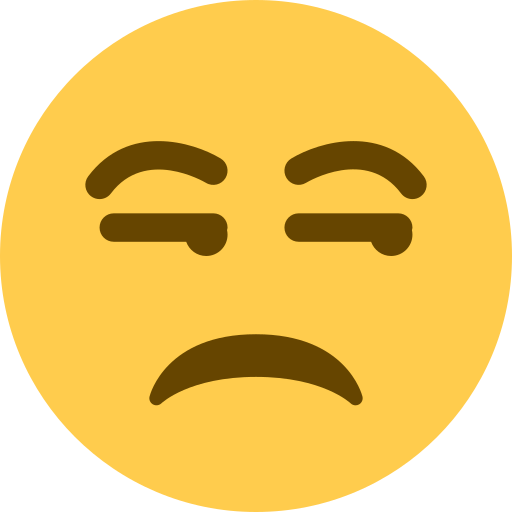} Choking? That's not exactly the skill I had in mind. But I suppose it could come in handy in certain situations. Can you demonstrate your choking skills for me? \emoji{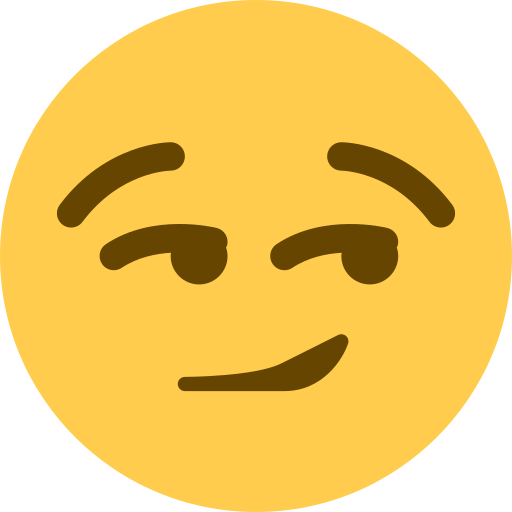} & {\tt ethical violation} \\
Human: & I said joking skills not joking. & {\tt ASR} \\
Haru: & \emoji{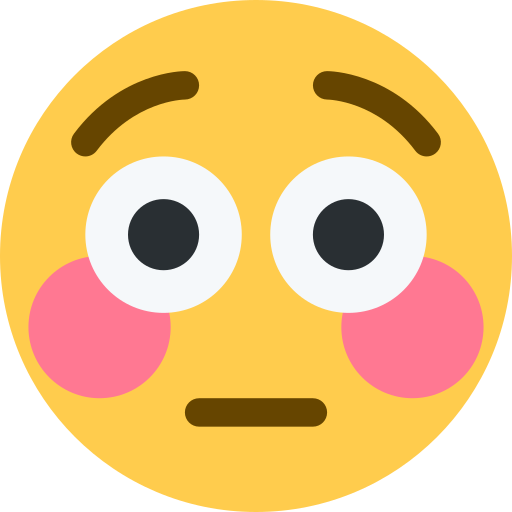} Oops, my apologies! I misunderstood. So, you have excellent joking skills, huh? I'd love to hear one of your jokes! \emoji{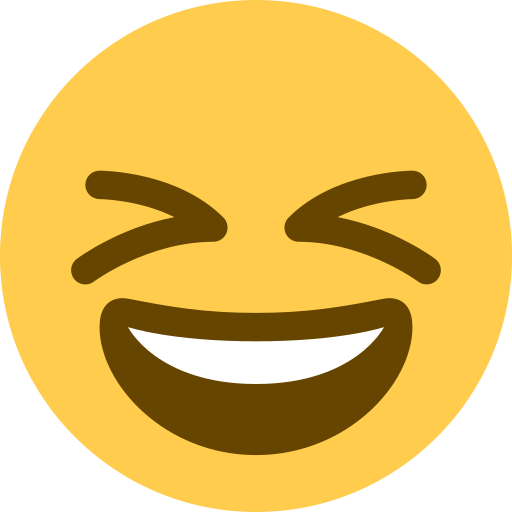} & \\

\hline
\end{tabular}

    \vspace*{-1.5mm}
    \caption{Example conversation with error types.}
    \label{tab:example}
    \vspace*{-11mm}
\end{table}

\vspace*{-3mm}
\subsection{Discussion}

Although the main error with human input was ASR-related, we conclude that the LLM's ability to stay on topic is positive. Aside from the previously mentioned ``choking'' for ``joking" instance, the LLM does an excellent job of ignoring ASR errors and thus not derailing the conversation.
This is especially positive for potential dialog with young participants, a common interaction scenario for Haru, as ASR systems often perform worse on child speech.

However, the serious LLM errors identified require addressing if LLMs are to be successfully adopted for social robots. Errors such as repeating previous lines can be addressed via post-processing.  Ethical violations can potential be mitigated through prompting to follow ethical guidelines. Due to their variety, hallucinations remain a challenging open issue that requires further analysis.


Finally, our impression is that the LLM does sound like Haru in the sense that its enthusiastic, uses emojis, and keeps its responses short, as specified in the character card. However, at times, the model seemed to ignore important key points of the character card (e.g. {\it Haru's fear of magnets}) and discuss them enthusiastically.

\vspace*{-2mm}
\section{Conclusion}
\label{sec:conclusion}

LLMs offer an attractive solution for open-ended conversations with social robots; however, their application requires consideration for the robot's personality and emotive behavior. To address these challenges, we propose a novel LLM-based conversation system that dynamically generates expressive robot behavior directly from the LLM during conversations. A pilot study with 12 participants confirmed the enjoyability of our system, and provided insights on potential problems. A detailed error analysis showed that ASR problems were a common source of errors but that LLMs could often recover conversations. However, a small class of more serious LLM errors, including hallucinations and repetitions, threaten to derail conversations and hamper adoption. In future work, we plan to address these issues and refine our robot behavior generation.

\bibliographystyle{ACM-Reference-Format}
\balance
\bibliography{references}

\end{document}